\documentclass[10pt,twocolumn,letterpaper]{article}

\usepackage{cvpr}
\usepackage{times}
\usepackage{epsfig}
\usepackage{epstopdf}
\usepackage{graphicx}
\usepackage{amsmath}
\usepackage{amssymb}
\usepackage{graphicx}
\usepackage{times}
\usepackage{epsfig}
\usepackage{graphicx}
\usepackage{amssymb}
\usepackage{amsmath}
\usepackage{url}
\usepackage{breqn}
\usepackage{booktabs}
\usepackage{tabularx}
\usepackage{multirow}
\usepackage{algorithm}
\usepackage{algorithmic}
\usepackage{xcolor,colortbl}
\usepackage{cite}
\definecolor{Gray}{gray}{0.85}
\newcolumntype{a}{>{\columncolor{Gray}}c}

\usepackage[pagebackref=true,breaklinks=true,letterpaper=true,colorlinks,bookmarks=false]{hyperref}

\cvprfinalcopy 

\def\cvprPaperID{7306} 

\ifcvprfinal\fi
\begin{document}

\title{Over-crowdedness Alert! Forecasting the Future Crowd Distribution}

\author{
Yuzhen~Niu \and
Weifeng~Shi
\and
Wenxi~Liu
\and
Shengfeng~He
\and
Jia Pan
\and
Antoni B. Chan
\\
}

\maketitle

\begin{abstract}
   In recent years, vision-based crowd analysis has been studied extensively due to its practical applications in real world. In this paper, we formulate a novel crowd analysis problem, in which we aim to predict the crowd distribution in the near future given sequential frames of a crowd video without any identity annotations. Studying this research problem will benefit applications concerned with forecasting crowd dynamics. To solve this problem, we propose a global-residual two-stream recurrent network, which leverages the consecutive crowd video frames as inputs and their corresponding density maps as auxiliary information to predict the future crowd distribution. Moreover, to strengthen the capability of our network, we synthesize scene-specific crowd density maps using simulated data for pretraining. Finally, we demonstrate that our framework is able to predict the crowd distribution for different crowd scenarios and we delve into applications including predicting future crowd count, forecasting high-density region, etc. 
\end{abstract}

\section{Introduction}

In recent years, vision-based crowd analysis has been extensively researched, due to its wide applications in crowd management, traffic control, urban planning, and surveillance. The recent researches mainly focus on crowd counting \cite{zhang2016single,chan2008privacy,idrees2013multi,zhang2015cross}, multi-target tracking \cite{pellegrini2009you,Sadeghian2017Tracking}, motion pattern analysis~\cite{zhou2012coherent,yi2015understanding}, holistic crowd evaluation~\cite{zhou2013measuring}, crowd attribute learning~\cite{yi2014l0,shao2015deeply}, and pedestrian path prediction~\cite{alahi2016social,gupta2018social} in images or videos. 

In real-world scenarios, in order to manage crowd behavior, it is critical to forecast the dynamics of crowd motion to prevent the dangers brought by over-crowded people, such as crowd crush that may cause people falls or fatalities. 
Existing research either investigate the previous or current status of the crowd \cite{zhang2016single,chan2008privacy,idrees2013multi}, or predict the individual trajectories within a less crowded scene ~\cite{alahi2016social,gupta2018social}. These methods can hardly be applied in situations to issue an alert for the potential danger of large-scale crowd in advance. On the other hand, little attention has been paid to predict the dynamics for large crowds holistically in the short-term or long-term futures. 

\begin{figure}
	\hspace{-0.3mm}
		\includegraphics[width=0.5\textwidth]
		{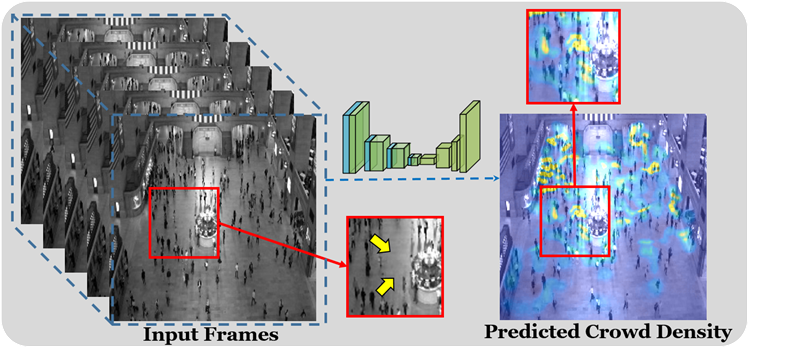}
	\vspace{-5mm}
	\caption{We formulate a novel problem to forecast the crowd distribution from sparsely sampled previous crowd video frames without knowing the individual identities. To solve this problem, we propose a prediction model that is able to learn the crowd dynamics to predict the crowd density in the near future. As illustrated, observing the crowd gathering behavior within the red box, our model manages to forecast the high-density area (indicated by the yellow region) ahead of time.}
	\label{fig:teaser}
\end{figure}

Hence, in this paper, we formulate a novel yet challenging crowd distribution prediction problem. Given several sequential frames of a crowd video without any exact position or identity information of the individuals, our goal is to estimate the crowd distribution in the near future (see Fig.~\ref{fig:teaser}). To benefit long-term prediction, the provided frames of the crowd video are sampled sequentially yet sparsely over an equal interval (up to 6 seconds), and we aim to predict the crowd distribution of the very next frame in the same interval. 
Specifically, the reason of sampling input frames over a large interval is that it allows to observe more variations of crowd dynamics and inject contextual information for a longer-term prediction.
Compared with tracking and path prediction tasks, the challenges of our problem is the identities or the positions of pedestrians are not provided in the input. Although it  mitigates the laborious annotation efforts in real-world application scenarios, the difficulty of prediction is also increase.
Besides, instead of predicting trajectories as outputs, we forecast the crowd distribution in the form of future crowd density map, which is informative for analyzing crowd dynamics, monitoring the high-density regions, and even detecting the abnormal crowd behavior. 
Furthermore, enabling the crowd density prediction without revealing the identities can well preserve the privacy of individuals in certain applications.

To solve the posed challenge, we propose a global-residual two-stream network to forecast the crowd density given the input sequential frames of the crowd video. In the first stream, given the input frames, we adopt a multi-scale recurrent network which extracts spatial context feature and leverage a series of convolutional LSTM layers, or a ConvLSTM block, to correlate the spatial and temporal features. 
To enhance the prediction, in the second stream, we set up a recurrent auto-encoder to predict the future crowd density from the corresponding density maps of the given frames. Since the sequential density maps can provide more abstract representation of crowd status and dynamics, it enables the prediction of crowd dynamics to be more accurate. 
Moreover, to further strengthen the capability of the second stream, we simulate diverse crowd behaviors, and thus generate a large amount of synthetic crowd density maps for pretraining. 
The computed features will be jointly passed through an attention-based module to forecast the future crowd density.
Finally, to incorporate the recent motion prior into prediction, we introduce an additional branch that combines the warped density map guided by flow map with our fused feature so as to improve the quality of the predicted density map. 

In experiments, we adopt the public video-based crowd counting datasets, UCSD~\cite{chan2008privacy} and Mall \cite{chen2012feature}, to evaluate the crowd density prediction. However, existing crowd video sequences are often too short to observe the complex dynamics of crowd, or the captured crowd scenes under limited camera views lead to little variation of crowd density.
Therefore, we manually annotate the crowd from an over 30-min video~\cite{zhou2012understanding} captured with a large camera view in the Grand Central Station, New York. For evaluating the predicted density map, we propose a metric that hierarchically measures the difference of local crowd density between predictions and ground-truths. Besides, we perform comprehensive experiments to compare our approach with the optical flow-based methods and video frame prediction approaches. 

To sum up, the contributions of our paper are fourfold:
\begin{itemize}
	\item We formulate a novel problem for predicting crowd density in the near future, given the past sequential yet sparsely sampled crowd video frames. 
	\item We propose a global-residual two-stream network architecture that learns from the crowd videos and the corresponding density maps separately to forecast the future crowd density.
	\item We incorporate different motion priors into the density prediction by simulating diverse synthetic density maps. It largely enriches the feature representations and robustness of the network.
	\item For evaluation, we manually label a long duration and large scale crowd video and we propose a spatial-aware metric for measuring the quality of the predicted crowd density. Moreover, we delve into several related crowd analysis applications.
\end{itemize}

\section{Related Works}

In this paper, we propose a novel research problem, future crowd distribution prediction. In this section, we will survey related aspects of our work, including crowd counting, path prediction, and video frame prediction. 

\textbf{Crowd counting} has been studied for years in computer vision \cite{kang2019beyond}, whose purpose is to count the number of people and to estimate how crowd is spatially arranged in images. Detection- or tracking-based methods \cite{brostow2006unsupervised,wu2005detection,rabaud2006counting} can solve the counting problem, but their performance are often limited by low-resolution and severe occlusion. In recent years, regression-based methods have been investigated for counting~\cite{chen2013cumulative,idrees2013multi}. Specifically, they directly map the image features to the number of people, without explicit object detection, or map local features to crowd blob count based on segmentation~\cite{chan2008privacy}. Besides, the concept of \textit{density map}, where the integral (sum) over any sub-region equals the number of objects in that region, was first proposed in \cite{lempitsky2010learning}. The density values are estimated from low-level features, thus sharing the advantages of general regression-based methods, while also maintaining location information~\cite{lempitsky2010learning,arteta2014interactive}. With the progress of deep learning techniques, convolutional neural network (CNN)-based methods have demonstrated excellent performance on the task of counting dense crowds~\cite{cao2018scale,zhang2015cross,zhang2016single,sam2017switching,shi2018crowd,liu2019point,li2018csrnet,ma2019bayesian}. Most of these methods first estimate the density map via deep neural networks and then calculate the counts. 
Unlike prior works on crowd counting, our work aims at predicting the crowd spatial distribution in the near future, given the multiple previously observed crowd images.

\begin{figure*}
	\vspace{-0.5cm}
	\begin{center}
		\includegraphics[width=1.0\linewidth]
		{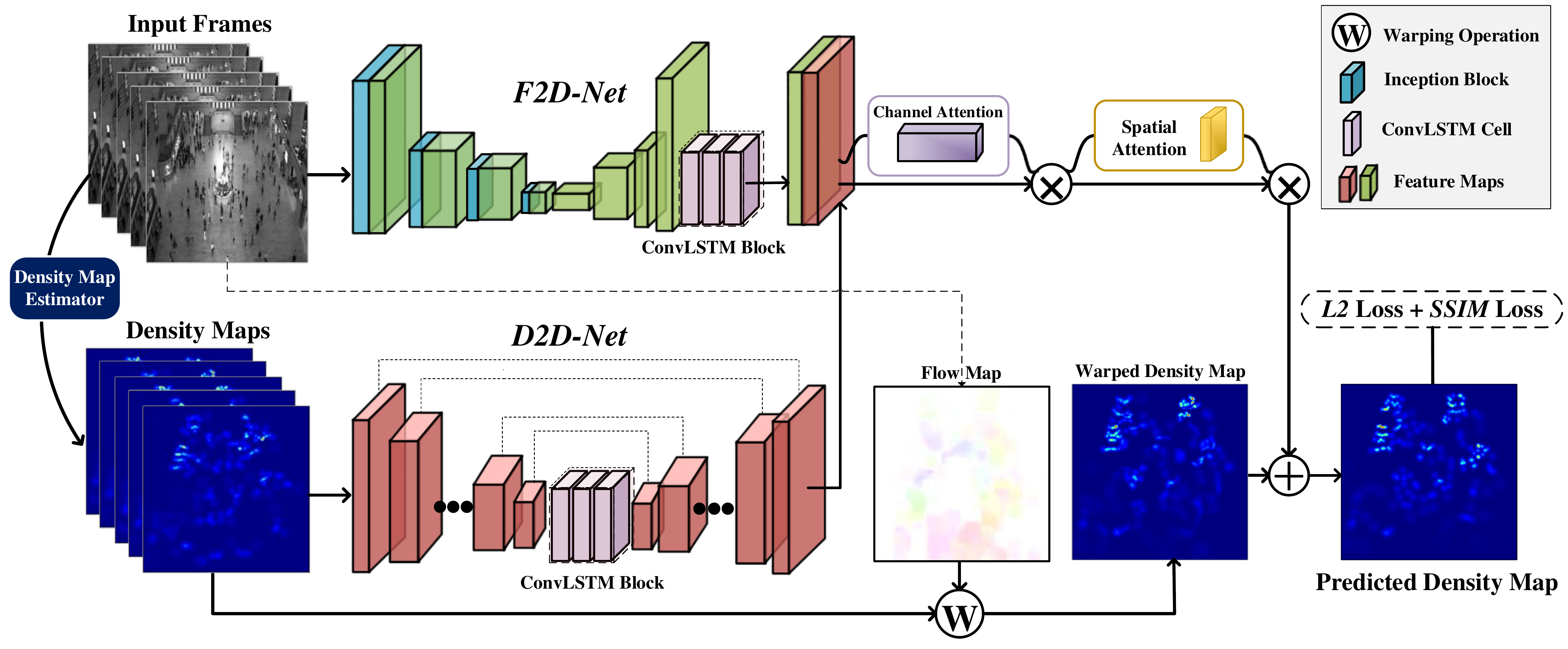}
	\end{center}
	\vspace{-0.3cm}
	\caption{Illustration of our network architecture. Our model is mainly composed of \textit{F2D-Net} and \textit{D2D-Net}. In particular, the input frames and their estimated density maps are separately fed into these two recurrent networks. After that, their output features are concatenated and passed through an attention-based fusion module. In the end, to strengthen the prediction, a global residual branch incorporates the motion information with the fused features into the final predicted density map.}
	\vspace{-3mm}
	\label{fig:network architecture}
\end{figure*}

\textbf{Trajectory prediction} is another related research topic that learns to forecast the human behavior under complex social interactions in the term of trajectory~\cite{yi2016pedestrian,alahi2016social,ma2017forecasting,bhattacharyya2018long}. In these methods, the focus is rested on human-human interaction, which has been investigated for decades in social science, graphics, vision, and robotics~\cite{helbing1995social,antonini2006discrete,van2008reciprocal,pellegrini2009you,liu2014leveraging,kim2015brvo,liu2016exemplar,long2017deep}. The interaction has been exhaustively addressed by traditional methods based on hand-crafted features~\cite{antonini2006discrete,yamaguchi2011you,pellegrini2009you}. Social awareness in multi-person scenes has been recently revisited with data-driven techniques based on deep neural networks~\cite{yi2016pedestrian,alahi2016social,gupta2018social,zhang2019sr,zhao2019multi,sadeghian2019sophie,makansi2019overcoming}. All these methods require the identities of the persons with their previous positions, and their studies are mostly evaluated on low-density or medium-density of crowd motions. Compared with them, our approach is able to work on large crowd scenes with a varying density without knowing the identities of individuals in the crowd. 

\textbf{Video frame prediction} recently achieves significant progress due to the success of Generative Adversarial Network (GAN)~\cite{Goodfellow2014GAN}. It is first studied to predict future frames for Atari game~\cite{Oh2015Action} and then researchers try to predict the future frames of natural videos~\cite{Liang2017Dual,Lu2017Flexible,Walker2017The,babaeizadeh2017stochastic,jayaraman2018time,liu2018future,lotter2016deep,mathieu2015deep}. In order to predict realistic pixel values in future frames, the model must be capable of capturing pixel-wise appearance and motion changes so as to let pixel values in previous frames flow into new frames. Different from these approaches, our prediction is based on sparsely sampled crowd video frames with the interval larger than 1.5 seconds. It is a much longer interval than the inputs of the video frame prediction methods, which brings challenges to our problem.

\section{Our Proposed Method}

In this section, we first present our problem formulation and then introduce our proposed network architecture. After that, we depict how to enhance our network by synthetic crowd data. 

\subsection{Problem Formulation}

\begin{figure*}
	\footnotesize
	\begin{center}
		\begin{tabular}{ccccccc}
			\includegraphics[width=0.12\textwidth]{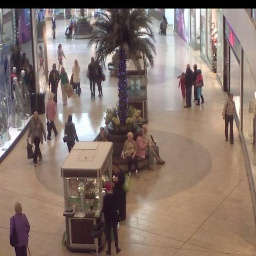}&
			\includegraphics[width=0.12\textwidth]{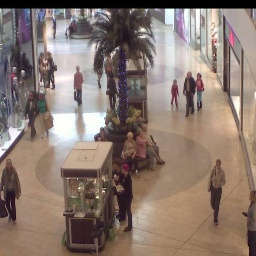}&
			\includegraphics[width=0.12\textwidth]{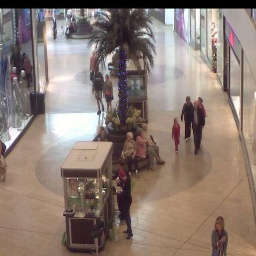}&
			\includegraphics[width=0.12\textwidth]{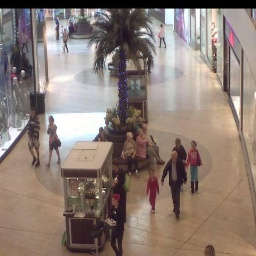}&
			\includegraphics[width=0.12\textwidth]{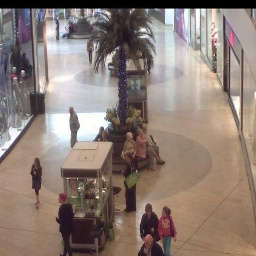}&
			\includegraphics[width=0.12\textwidth]{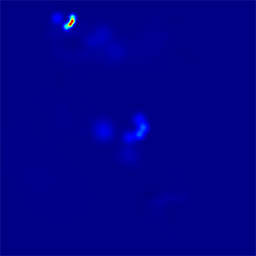}&
			\includegraphics[width=0.12\textwidth]{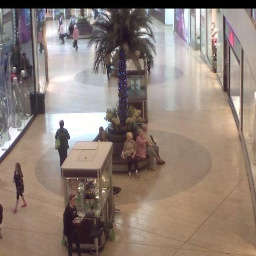}\\
			f = 1684 &  f = 1687 & f = 1690 & f = 1693 & f = 1696 & Pred. for f=1699 & f = 1699\\

			\includegraphics[width=0.12\textwidth]{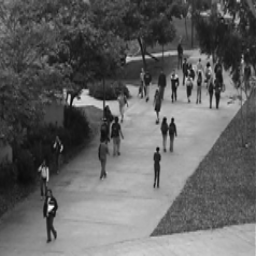}&
			\includegraphics[width=0.12\textwidth]{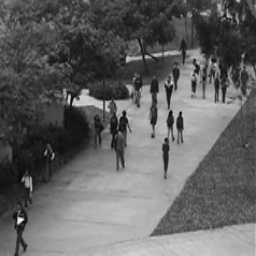}&
			\includegraphics[width=0.12\textwidth]{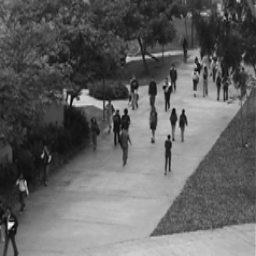}&
			\includegraphics[width=0.12\textwidth]{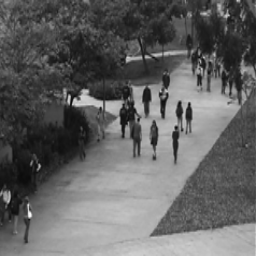}&
			\includegraphics[width=0.12\textwidth]{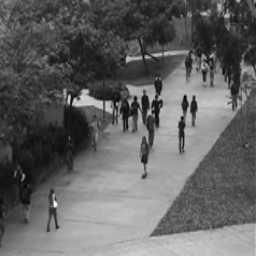}&
			\includegraphics[width=0.12\textwidth]{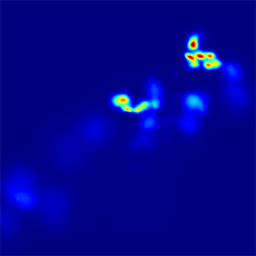}&
			\includegraphics[width=0.12\textwidth]{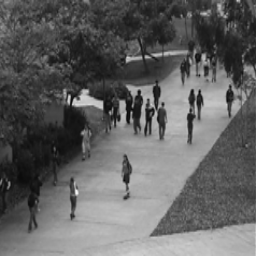}\\
			(a) f = 1661 & (b) f = 1676 & (c) f = 1681  & (d) f = 1696  & (e) f = 1711  & (f) Pred. for f=1726 & (g) f = 1726\\
		\end{tabular}
	\end{center}
    \vspace{-0.3cm}
	\caption{Examples of predicted density maps (pred.) from the Mall and UCSD datasets. The first five columns are the input frames to our model, which are sparsely sampled from the original video (the frame number is denoted as f). For Mall, the inputs are sampled every 3 frames ($\Delta t > 1.5$ seconds). For UCSD, the inputs are sampled every 5 frames ($\Delta t = 1.5$ seconds).}
	\vspace{-0.3cm}
	\label{fig:show}
\end{figure*}

In this paper, we introduce a novel research problem for crowd analysis. Given a sequence of crowd video frames, the goal is to predict the crowd distribution in the near future. For forecasting crowd dynamics, it is critical to predict the crowd status in a longer period of time, so that it may facilitate practical applications, e.g., issuing alerts for over-crowd situations beforehand.
To benefit the long-term prediction, the given crowd video frames are sampled at a certain equal interval (e.g. 1.5 seconds) and the task is to predict the crowd status at the very next time step. Hence, we can formulate it as: 
\begin{align}
	D_{t+N\Delta t} = \mathcal{F}(\{I_{t}, I_{t+\Delta t},\cdots, I_{t+(N-1)\Delta t}\}),
\end{align}
where the input frames of our model $\mathcal{F}$ are denoted as $\{I_{t}, I_{t+\Delta t},\cdots, I_{t+(N-1)\Delta t}\}$, which contains $N$ frames sequentially sampled from video with an equal interval $\Delta t$. Given the input frames, our model is required to predict the crowd density $D$ at the next time step $t+N\Delta t$. We show two crowd density prediction examples in Fig.~\ref{fig:show} from the Mall and UCSD datasets. 

\subsection{Network architecture}

As illustrated in Fig. \ref{fig:network architecture}, we propose a global-residual two-stream network for predicting crowd density. In general, our framework consists of several main modules: the \textit{Frame-to-Density} network (i.e., \textit{F2D-Net} that is able to predict density from sequential crowd video frames),
the \textit{Density-to-Density} network (i.e., \textit{D2D-Net} that predicts future density from sequential density maps), the density map estimator that estimates the crowd density from a single crowd image, the attention-based feature fusion module, and a global-residual branch based on the warped density map estimated from flow map of the input video frames.

\textbf{F2D-Net.} As the first stream of our framework, F2D network, fed with the frames from video, is composed of a multi-scale convolutional blocks for extracting spatial feature from the input frames and a series of convolutional LSTM cell, or a ConvLSTM module to learn the spatial-temporal correlation from sequential data. As shown in Fig.~\ref{fig:network architecture}, we adopt several inception blocks that contain four subbranches with filter size of $1 \times 1$, $3 \times 3$, $5 \times 5$, and $7 \times 7$ for extracting multi-scale features. Then, the feature maps are passed through convolutional layers and the transposed convolutional layers to further transform the spatial features and upsample their spatial dimension.

In addition, to better model the spatio-temporal feature in the frames, we incorporate a ConvLSTM block at the end of F2D network, which is made up of a $3 \times 3$ ConvLSTM cell with filter size 16 and two $1 \times 1$ ConvLSTM cells with filter size 16 and 1, respectively. Specifically, the ConvLSTM cell is a dominant recurrent layer that can capture the spatio-temporal correlations from the sequential data and preserve the dimension of the output as the same size of the input. 

\textbf{D2D-Net.} Since the input frames are sampled from a crowd video sparsely, we want to introduce auxiliary information to assist the prediction of crowd density in the near future. Thus, in the second stream of the framework, we input the corresponding density maps of the given video frames for enhancing the temporal prediction. The density maps provide more abstract representation than the video frames, so it can make the model more robust. As shown in Fig.~\ref{fig:network architecture}, to predict crowd density from density maps of the past frames, we set up a recurrent encoder-decoder structure similar to U-Net~\cite{ronneberger2015u} consisting of downsampling and upsampling stages. In the downsampling stage, we use eight $3 \times 3$ convolutional layers to extract the compact features from the density map sequences. To bridge the downsampling and upsampling stages, we apply another ConvLSTM block, which is composed of three ConvLSTM cells with filter size $5$ to better capture the spatio-temporal correlations. Besides, in the upsampling stage, 
it consist of eight upsampling layers to generate the output with the same size as the input density map. In addition, the skip connections are used to combine low-level details to the high-level semantics between the downsampling and upsampling stages. 

\textbf{Density map estimator.} To obtain the corresponding density map of each input frame, we adopt the crowd counting model \cite{cao2018scale} that is able to regress a single crowd image to a density map.

\textbf{Attention-based fusion module.} We concatenate the outputs of F2D network and D2D network, as shown in Fig. \ref{fig:network architecture}, and then pass it into the attention-based fusion module. Specifically, we sequentially incorporate a channel-wise attention module and a spatial attention module. 

\begin{figure}
	\footnotesize
	\begin{tabular}{ccc}
		\includegraphics[width=0.15\textwidth]{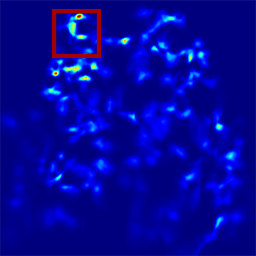}&
		\includegraphics[width=0.15\textwidth]{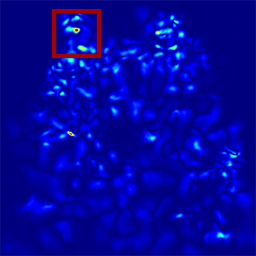}&
		\includegraphics[width=0.15\textwidth]{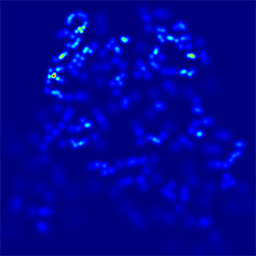}\\
		(a) Flow map & (b) Residual & (c) Ground truth \\
	\end{tabular}
	\caption{We demonstrate an example of the input flow map and the residual learned by our two-stream network. The residual map compensates and refines the inaccurate motion information brought by the flow map. }
	\vspace{-3mm}
	\label{fig:residual}
\end{figure}

\textbf{Global residual branch.} Since we aim at predicting the crowd status in the near future, the most recent motion information will be most reliable for improving the prediction. To incorporate the motion prior, we calculate the optical flow by \cite{IMKDB17} and propagate the latest density map based on the computed flow to obtain the warped density map, as illustrated in Fig. \ref{fig:network architecture}. Thus, our fused features will join the warped density as the global residual to generate the future density (see Fig.~\ref{fig:residual}). 
At the end, we adopt a hybrid of SSIM~\cite{ssim} and $L2$ loss as the training loss.

\subsection{Synthetic crowd data}

In our framework, D2D-Net serves as important auxiliary information to improve the prediction. But, as mentioned, the input sequence of D2D-Net is the estimated density data of the crowd video frames, which normally will not bring in the new information for the framework. In practice, it is difficult to access or annotate a large amount of crowd density data for training. To mitigate this problem and thoroughly strengthen the ability of the D2D-Net, we propose to pretrain D2D-Net alone using scene-specific crowd simulation data. 

\begin{figure}
	
	\footnotesize
	\begin{tabular}{ccc}
		\includegraphics[width=0.21\textwidth]{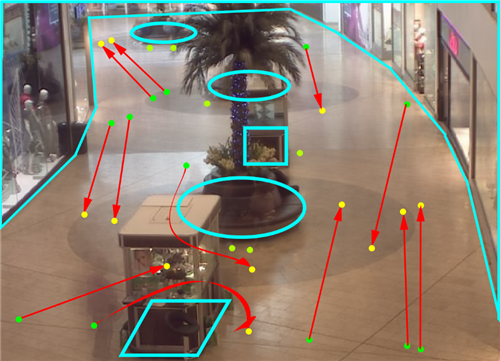}&
		\includegraphics[width=0.12\textwidth]{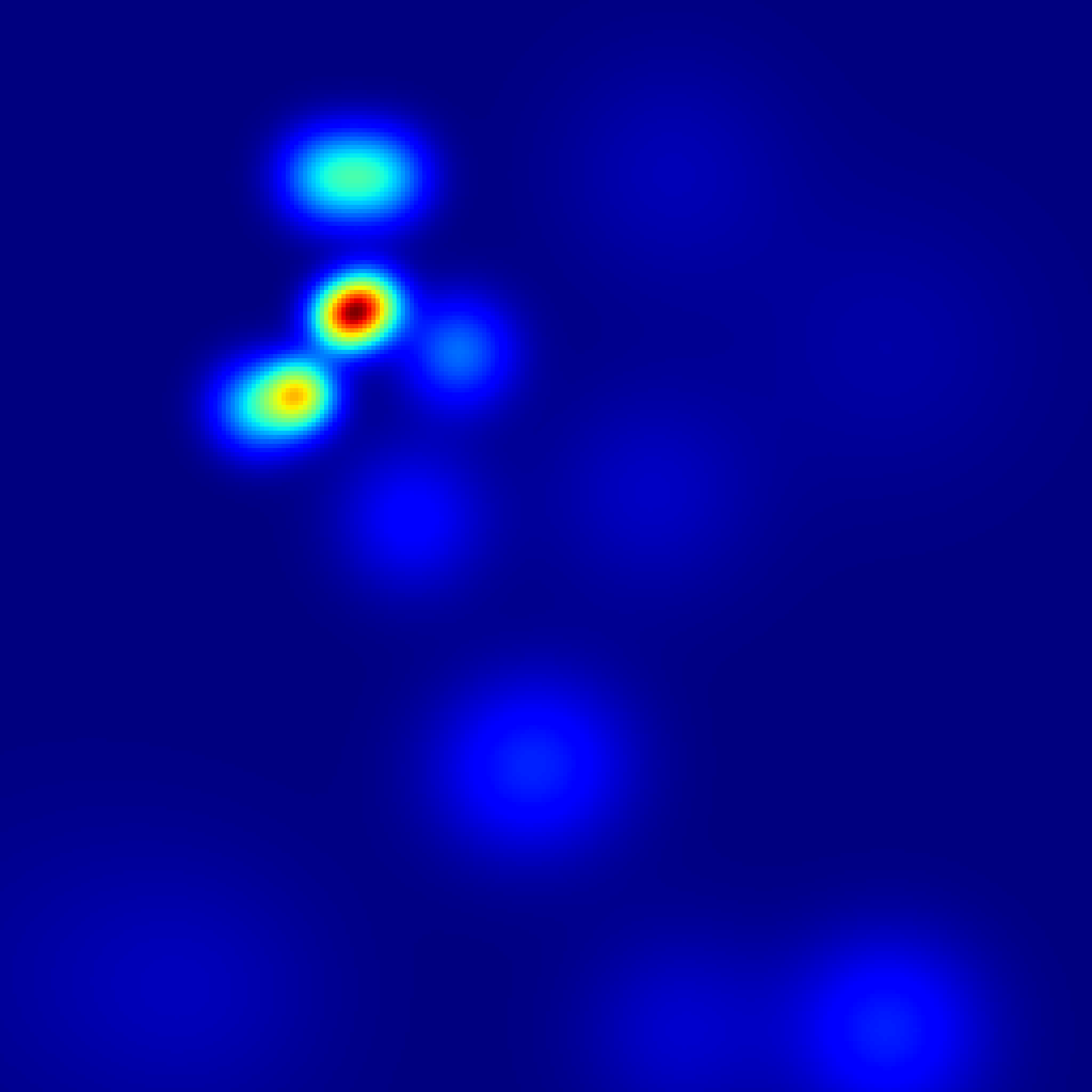}&
		\includegraphics[width=0.12\textwidth]{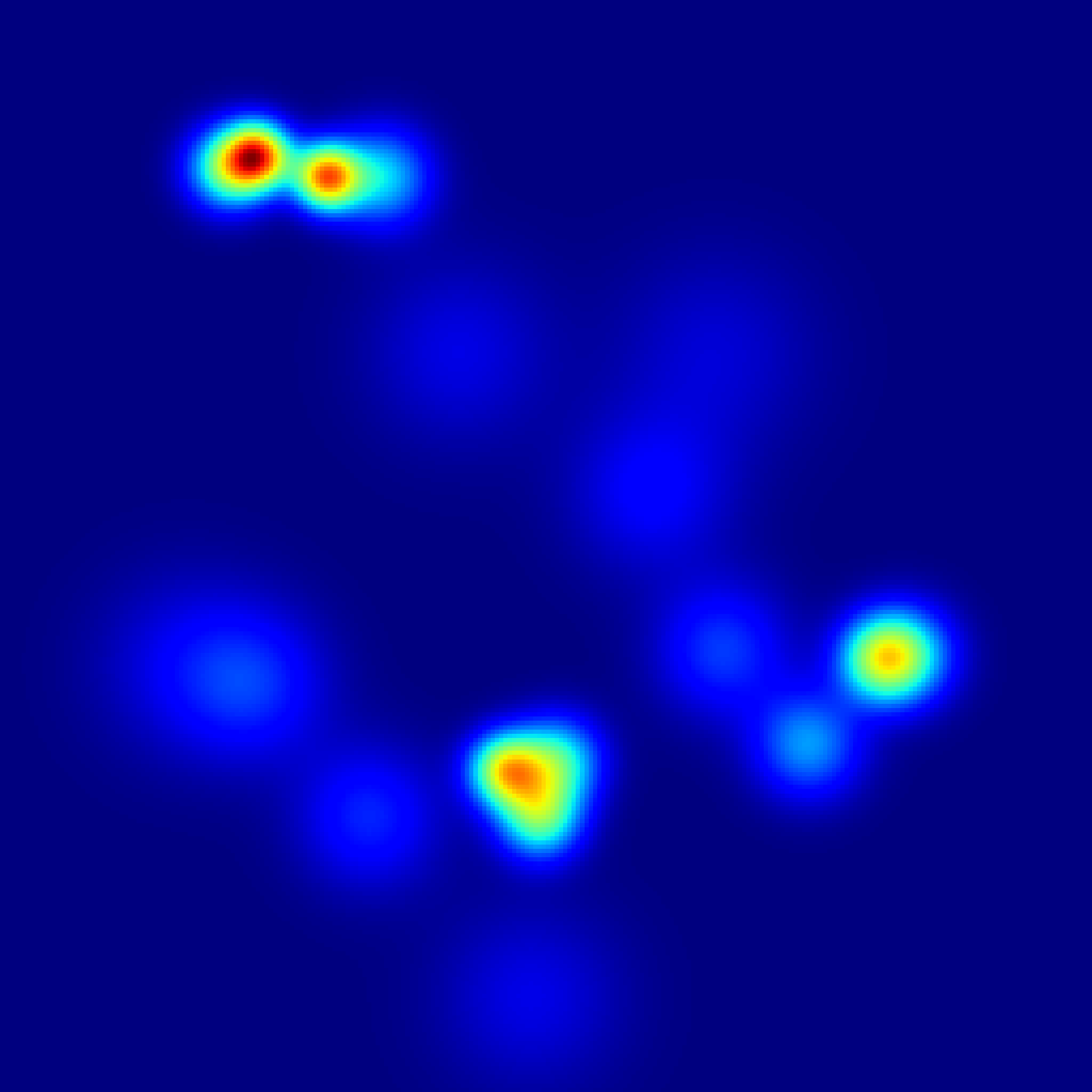}\\
		(a) Synthetic crowd motion & (b) $\tilde{D_{t}}$ & (c) $\tilde{D}_{t+\Delta t}$
	\end{tabular}
	\caption{Example of synthetic crowd data (Best viewed in color). In (a), we show an example of simulated crowd motion from the time step $t$ to $t+\Delta t$, where the green and yellow dots represent the starting and ending positions of each person, while red arrows indicate their motion directions. The cyan shapes are obstacles in the ground space of the scene. Given the simulated data, we can synthesize the density map before the crowd moves, $\tilde{D_{t}}$, and the one after the crowd moves, $\tilde{D}_{t+\Delta t}$, in a period of $\Delta t$. }
	\label{fig:sim}
\end{figure}

Crowd simulator has been studied for decades in the research area of graphics and robotics, which can be used to model collision-free dense crowd behavior. Here, we adopt a well-known crowd simulator~\cite{helbing1995social} to generate diverse crowd dynamics, by randomly initializing the starting states and the destinations for pedestrians. As show in Fig. \ref{fig:sim}, after obtaining a simulated crowd motion, we can project the individual positions from the world space (or the ground space) of the simulation to the image space using homography transformation. 
Then, to generate the density map from the calculated individual positions, we follow \cite{zhang2016single} and use the geometry-adaptive kernel to generate the density map. Since we require the density maps only, it will not be a strict demand for the realism of the crowd simulation.
In this way, the simulated data can be converted to synthetic density maps, and thus can be applied for pretraining our D2D-Net.

\section{Experimental Results}

\subsection{Datasets and implementation details}

\begin{table}
	\centering
	\footnotesize
	\caption{Statistics of crowd density prediction benchmarks.}
	\begin{tabular}{c||ccc}
		\toprule
		Dataset & UCSD  & Mall  &  Station \\
		\midrule
		Duration (\textit{sec.}) & 200 & $>$ 1000 & 2000 \\
		Annotated frame rate & 10 & $<$ 2 & 0.5 \\
		\# of persons per frame & 24.9 & 31.2 & 187.7 \\
		Min \# of persons per frame & 11 & 13 & 116 \\
		Max \# of persons per frame & 46 & 53 & 374\\
		STD of persons per frame & 9.6 & 6.9 & 37.6\\
		Total \# of annotated persons & 49,885 & 62,315 & 187,744\\
		\bottomrule
	\end{tabular}
	\label{tab:dataset}
\end{table}

\textbf{Datasets.} To evaluate our approach, our experiments are conducted on two public datasets: \textit{UCSD}~\cite{chan2008privacy} and \textit{Mall} \cite{chen2012feature}. In addition to these datasets, we also manually annotate the pedestrian distributions from a 33-minute crowd video~\cite{zhou2012understanding} which is captured in the Grand Central Station, denoted as \textit{Station}. The statistics of these datasets are provided in Table \ref{tab:dataset}. Compared with \textit{UCSD} and \textit{Mall}, \textit{Station} contains more persons and larger scenarios, leading to more complex crowd distribution and behaviors. Besides, the duration of \textit{Station} is much longer than the other two datasets and thus it can be observed that the crowd density obviously varies over time, which is more suitable for evaluating our method. 

\textbf{Implementation details.} Our framework is implemented using Tensorflow, which is trained and tested on a PC with a Tesla P100 GPU. The input images are normalized to $256\times 256$ and our model outputs the density map with the same size. During training, our network is trained with a mini-batch size of 8. The initial learning rate is set to $1 \times 10^{-4}$ and the exponential decay is applied to the learning rate for every 1,000 training steps. 

\subsection{Metric}

To evaluate how well the density can be predicted, we adopt the following metrics. 
Since we are interested in the global and local prediction performance, we divide the predicted density map into equal $K$ patches and compute the local count for each patch. Following the standard practice in crowd counting~\cite{lempitsky2010learning}, we accumulate the values of each patch from the predicted density map to obtain the local counts.
Then, we measure them using their corresponding ground-truths to calculate the mean absolute error of prediction (P-MAE) and the mean square error of prediction (P-MSE), which is measured as:
\begin{align}
	\text{P-MAE}_K = \frac{1}{N \cdot K}\sum_{n=1}^N\sum_{k=1}^K \|C^n_k - \sum_{(i,j)\in D^n_k} D^n_k[i,j]\|_1,
	\nonumber\\
	\text{P-MSE}_K = \frac{1}{N \cdot K}\sum_{n=1}^N\sum_{k=1}^K \|C_k - \sum_{(i,j)\in D_k} D_k[i,j]\|^2_2, \nonumber
\end{align}
where $N$ denotes the total number of predicted density maps in test. $D^n_k$ refers to the $k^{th}$ patch within the $n^{th}$ predicted density map $D^n$ and $C^n_k$ represents the ground-truth count of the $k^{th}$ patch. 
In experiments, we choose $K=1, 4, 16$, and $64$. When $K=1$, the metric is actually adopted to measure the global density map. 
Thus, the patch sizes are $256 \times 256$, $128 \times 128$, $64 \times 64$, and $32 \times 32$,  respectively.


\subsection{Ablation study}

In this subsection, we first analyze the structure of our proposed model. 
We perform the ablation experiments based on the \textit{Mall} dataset using the metrics mentioned above.
First of all, we compare the performance of the following structures: (1) \textit{F2D-Net} trained using crowd video frames; (2) \textit{D2D-Net} trained on ground-truth density maps; (3) the joint network of \textit{F2D-Net} and \textit{D2D-Net}, denoted as \textit{Joint}, trained on video frames and the corresponding ground-truth density maps; (4) the joint network whose \textit{D2D-Net} is pretrained on synthetic crowd data and then trained on ground-truth density maps, denoted as \textit{Joint-Sim}; (5) the joint network is same as (4), but the density maps are unknown and thus estimated by the density estimator during test; (6) the joint network with optical flow as input, denoted as \textit{Joint-Flow}; (7) the network is the same as (6), except that the density maps are unknown during test. The comparison results are demonstrated in Table \ref{tab:ablation1}. 

\begin{figure}
	\vspace{-0.4cm}
	\footnotesize
	\begin{tabular}{ccc}
		\includegraphics[width=0.15\textwidth]{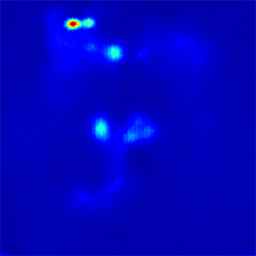}&
		\includegraphics[width=0.15\textwidth]{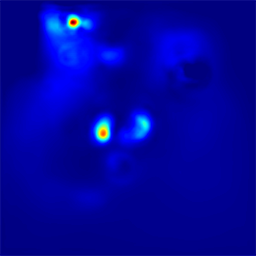}&
		\includegraphics[width=0.15\textwidth]{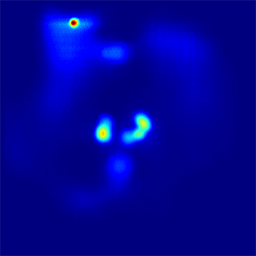}\\
		(a) \textit{F2D-Net} & (b) \textit{D2D-Net} & (c) \textit{Joint} \\
		P-MAE: 0.370 &  P-MAE: 0.299 & P-MAE: 0.289 \\
		\includegraphics[width=0.15\textwidth]{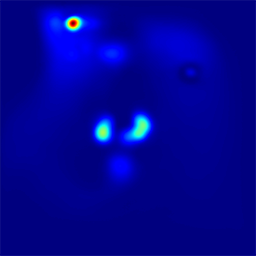}&
		\includegraphics[width=0.15\textwidth]{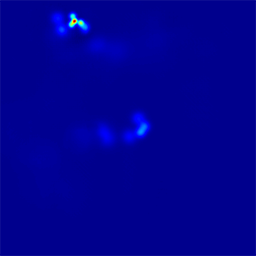}&
		\includegraphics[width=0.15\textwidth]{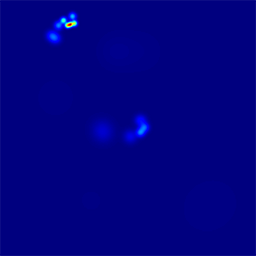}\\
		(e) \textit{Joint-Sim} & (f) \textit{Joint-Flow} & (g) Ground truth\\
		P-MAE: 0.287 &  P-MAE: 0.168 & 
	\end{tabular}
	\caption{An example of crowd density prediction from Mall to demonstrate the effectiveness of different modules (a-f) with the reference of ground-truth (g). The warmer color indicates the higher density, and vice versa. We also show their respective P-MAE$_K$ with K=64.}
	\label{fig:ablation1}
\end{figure}

According to the results measured in P-MSE, the joint network of \textit{F2D-Net} and \textit{D2D-Net} increases the performance of the separate networks. With the addition of the synthetic data and the motion information introduced by flow map, the model can further be improved. Note that, without the ground-truth density maps as input, the performance of the model will slightly decrease. Furthermore, we also illustrate an example selected from the Mall dataset in Fig.~\ref{fig:ablation1}. The results are consistent with Table \ref{tab:ablation1}.
In particular, in Fig.~\ref{fig:ablation1}(a), \textit{F2D-Net} produces blurred results, due to the inaccurate frame-based prediction. \textit{D2D-Net} learns temporal transitions of the density maps, which roughly localizes the high-density areas. Combining these two networks, the prediction result of the joint network (c) is improved. It can be slightly enhanced by pretrained synthetic data. Given the flow map that introduces the motion information, the result becomes more accurate with the reference of GT.

\begin{table}
	\vspace{-0.4cm}
	\begin{center}
		\footnotesize
		\setlength{\tabcolsep}{7pt}
		\caption{Ablation studies on the network structures. GT-D refers to the model inference is performed with or without the ground-truth density maps as auxiliary input.}
		\begin{tabular}{cc||cccc}
			\toprule
			\multicolumn{2}{c||}{Metric} & \multicolumn{4}{c}{P-MSE} \\
			\midrule
			Structure & GT-D & K=1 & K=4 & K=16 & K=64\\
			\midrule
			\textit{F2D-Net} & $\checkmark$ & 35.567 & 8.632  & 1.767 & 0.360 \\
			\textit{D2D-Net} & $\checkmark$ & 41.011 & 10.491 & 2.023 & 0.377 \\
			\textit{Joint} & $\checkmark$ & 30.987 & 9.301  & 1.845 & 0.360\\
			\midrule
			\multirow{2}{*}{\textit{Joint-Sim}} & $\checkmark$ & 27.039 & 7.586  & 1.604 & 0.321\\
			 & $\times$ & 33.202 & 9.254  & 1.867 & 0.362\\	
			\midrule
			\multirow{2}{*}{\textit{Joint-Flow}} & $\checkmark$ & 17.884 & 4.617  & 1.248 & 0.281\\			
			& $\times$ & 20.113 & 5.309  & 1.415 & 0.311 \\			
			\bottomrule
		\end{tabular}
		\label{tab:ablation1}
		\vspace{-0.3cm}
	\end{center}
\end{table}

\begin{table}
    \vspace{-0.3cm}
	\begin{center}
		\footnotesize
		\setlength{\tabcolsep}{7pt}
		\caption{Varying numbers of input frames for our prediction.}
		\begin{tabular}{cc||cccc}
			\toprule
			\multicolumn{2}{c||}{Metrics} & \multicolumn{4}{c}{P-MSE}\\
			
			\midrule
			Method & Frames & K=1 & K=4 & K=16 & K=64 \\
			\midrule
			Flow & 2 & 18.098 & 5.429 & 1.564 & 0.350 \\\midrule
			 & 2 & 18.133 & 5.454 & 1.559 & 0.349 \\
			Ours & 3 & 17.635 & 5.343 & 1.547 & 0.347 \\
			 & 5 & 17.884 & 4.617 & 1.248 & 0.281 \\			
			\bottomrule
		\end{tabular}
		\label{tab:ablation2}
		\vspace{-0.3cm}
	\end{center}
\end{table}

%

Second, we evaluate the number of input frames. The default setting for our model is 5 frames. Here, we also validate it with 2 and 3 frames, respectively. For reference, we also compare with the flow based method that propagates the ground-truth density of the last input frame by optical flow. The results are depicted in Table \ref{tab:ablation2}. Compared with the flow-based estimation results, the model with 2 frames as input has similar performance. According to the results, more input frames lead to better performance, since they bring in more knowledge of crowd dynamics. 
Based on our observation, with more than 5 frames as input, it will not gain much improvements on performance, but it requires more computational resource and collecting more data for processing. 
Thus, in the following experiments, we use 5 input frames for comparison.

\begin{table*}
	\vspace{-0.4cm}
	\begin{center}
		\small
		\setlength{\tabcolsep}{7pt}
		\caption{Comparison results of different prediction methods in three benchmarks.}
		\vspace{-0.2cm}
		\begin{tabular}{cccc||cccc|cccc}
			\toprule
			\multicolumn{4}{c||}{Metric} & \multicolumn{4}{c}{P-MAE$\downarrow$} & \multicolumn{4}{|c}{P-MSE$\downarrow$} \\
			\midrule
			Dataset & $\Delta t$ (sec.) & Method & GT-D  & K=1 & K=4 & K=16 & K=64 & K=1 & K=4 & K=16 & K=64 \\
			\midrule
			\multirow{11}{*}{Mall}& & \textit{Flow+density} & $\checkmark$ & 3.256 & 1.741 & 0.831 & 0.304 & 18.098 & 5.429 & 1.564 & 0.350 \\
			& & Ours & $\checkmark$ & \textbf{3.189}  & \textbf{1.621} & \textbf{0.762} & \textbf{0.284} & \textbf{17.884}  & \textbf{4.617}  & \textbf{1.248} & \textbf{0.281} \\
			\cmidrule{3-12} 
			& $>1.5$ & VP & $\times$ & 14.752 & 3.841 & 1.112 & 0.338 & 239.884 & 26.590 & 2.944 & 0.468\\
			& & \textit{Flow+frame} & $\times$ & 3.582 & 2.117 & 1.011 & 0.365 & 20.299 & 7.479 & 2.005 & 0.406\\
			& & \textit{Flow+density} & $\times$ & 3.717 & 1.895 & 0.894 & 0.332 & 22.256 & 6.164 & 1.737 & 0.380 \\
			& & Ours & $\times$ & \textbf{3.522} & \textbf{1.757} & \textbf{0.824} & \textbf{0.310} & \textbf{20.113} & \textbf{5.309} & \textbf{1.415} & \textbf{0.311} \\ 
			\cmidrule{2-12} 
			 & & \textit{Flow+density} & $\checkmark$ & 4.425 & 2.427 & 1.081 & 0.375 & 30.618 & 10.654 & 2.574 & 0.521 \\
			& & Ours & $\checkmark$ & \textbf{4.062} & \textbf{2.121} & \textbf{0.960} & \textbf{0.349} & \textbf{25.988} & \textbf{7.983}  & \textbf{1.981} & \textbf{0.414} \\
			\cmidrule{3-12} 
			 & $>4.5$& \textit{Flow+frame} & $\times$ &5.424 & 2.810 & 1.271 & 0.435 & 45.766 & 13.570 & 3.027 & 0.571 \\
			 & & \textit{Flow+density} & $\times$ & 4.764 & 2.470 & 1.087 & 0.385 & 34.841 & 10.555 & 2.539 & 0.518 \\
			 & & Ours & $\times$ & \textbf{4.260} & \textbf{2.105} & \textbf{0.971} & \textbf{0.355} & \textbf{28.540} & \textbf{7.728}  & \textbf{1.930} & \textbf{0.412}  \\
			\midrule
			\multirow{6}{*}{UCSD} & & \textit{Flow+density} & $\checkmark$ & 1.506 & 0.827 & \textbf{0.369} & \textbf{0.150} & 4.171 & 1.143 & 0.351 & \textbf{0.096} \\
			& & Ours & $\checkmark$ & \textbf{1.481} & \textbf{0.791} & 0.374 & 0.153 & \textbf{3.592} & \textbf{1.092} & \textbf{0.344} & \textbf{0.096}\\
			\cmidrule{3-12} 
			& $1.5$ & VP & $\times$&17.115 & 4.298 & 1.097 & 0.287 & 299.067 & 34.044 & 3.267 & 0.396\\
			& & \textit{Flow+frame} & $\times$ &3.636 & 1.301 & 0.473 & 0.183 & 16.458 & 3.322 & 0.578 & 0.139 \\
			& & \textit{Flow+density} & $\times$  & 3.127 & 1.171 & 0.441 & \textbf{0.175} & 13.629 & 2.426 & 0.496 & 0.125 \\
			& & Ours & $\times$ & \textbf{2.266} & \textbf{1.049} & \textbf{0.436} & {0.178} & \textbf{7.984}  & \textbf{2.020} & \textbf{0.462} & \textbf{0.123} \\
			\midrule
			& & \textit{Flow+density} & $\checkmark$ & 8.921 & 4.171 & 2.002 & 1.018 & \textbf{408.872} & 44.889 & 8.295 & 2.188\\
			& & Ours & $\checkmark$ & \textbf{8.912} & \textbf{4.144} &\textbf{ 1.985} & \textbf{1.004} & 411.757 & \textbf{44.792} & \textbf{8.146} & \textbf{2.091}\\
			\cmidrule{3-12}
			 & $2$& VP & $\times$ & 122.140 & 30.535 & 7.681 & 2.032 & 15413.3 & 992.284 & 84.959 & 7.740\\
			& &  \textit{Flow+frame} & $\times$ & 16.276 & 6.022 & 2.671 & 1.184 & 578.549 & 68.682 & 13.059 & 2.888\\
			& &  \textit{Flow+density} & $\times$ & \textbf{14.328}  & 5.297  & 2.362 & 1.124 & \textbf{430.532}   & 53.464  & \textbf{10.231} & 2.650\\
			Station & & Ours &$\times$ & 14.439  & \textbf{5.290}  & \textbf{2.357} & \textbf{1.105} & 435.254   & \textbf{53.211}  & 10.248 & \textbf{2.558} \\
			\cmidrule{2-12}
			& & \textit{Flow+density} & $\checkmark$ & \textbf{11.473} & 5.721 & 2.839 & 1.366 & \textbf{516.536} & 67.233 & 14.940 & 3.895\\
			& & Ours & $\checkmark$ & 11.676 & \textbf{5.615} & \textbf{2.690} & \textbf{1.243} & 527.893 & \textbf{65.678} & \textbf{13.474} & \textbf{3.223}\\
			\cmidrule{3-12}
			& 6&  \textit{Flow+frame} & $\times$ & 16.936 & 6.957 & 3.284 & 1.495 & 641.969 & 88.717 & 19.083 & 4.480\\
			& &  \textit{Flow+density} & $\times$ & 15.276 & 6.637 & 3.086 & 1.455 & 510.906 & 78.796  & 16.723 & 4.367\\
			& & Ours &$\times$ & \textbf{14.692} & \textbf{5.770} & \textbf{2.509} & \textbf{1.201} & \textbf{458.186} & \textbf{60.904}  & \textbf{11.451} & \textbf{2.957}\\
			\bottomrule
		\end{tabular}
	    \vspace{-0.2cm}
		\label{tab:cmp}
		\vspace{-0.4cm}
	\end{center}
\end{table*}

\subsection{Result analysis}

Since our formulated problem has not been studied, we compare our model against the variants of the related techniques: the flow based methods and the video frame prediction (\textit{VP}). Specifically, we implement two flow-based methods: one is to propagate and warp the ground-truth density map of the most recent frame by optical flow; the other is to directly propagate and warp the most recent frame by optical flow and apply the density map estimator to compute its density map. They are denoted as \textit{Flow+density} and \textit{Flow+frame}, respectively. For \textit{VP}, we apply \cite{mathieu2015deep} to predict the appearance of the next future frame and then apply the density map estimator to compute its density map as well.
We compare these methods on three benchmarks, as shown in Table~\ref{tab:cmp}. For a thorough comparison experiment, we compare them with different conditions: the interval ($\Delta t$) for the input frames and whether the ground-truth density maps of the input frames are known during the inference stage (i.e. GT-D). We measure the comparison using P-MAE and P-MSE in four scales, i.e. $K=1,4,16,64$, respectively.

\begin{figure*}
	\footnotesize
	\begin{center}
		\begin{tabular}{cccccc}
			\includegraphics[width=0.15\textwidth]{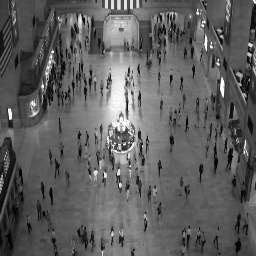}&
			\includegraphics[width=0.15\textwidth]{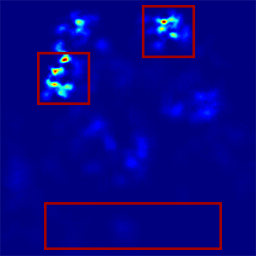}& 
			\includegraphics[width=0.15\textwidth]{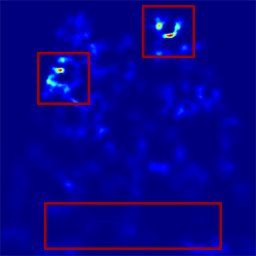}&
			\includegraphics[width=0.15\textwidth]{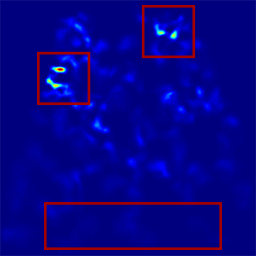}&
			\includegraphics[width=0.15\textwidth]{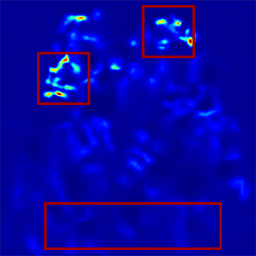}&
			\includegraphics[width=0.15\textwidth]{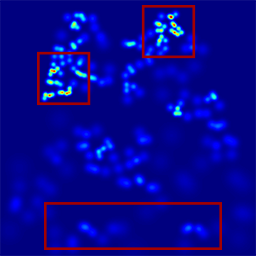}\\
			Future frame & VP & Flow + frame & Flow + density & Ours & GT density \\
			             & P-MAE: 1.957 &  P-MAE: 1.483 & P-MAE: 1.576 & P-MAE: 1.085 &  \\
			\includegraphics[width=0.15\textwidth]{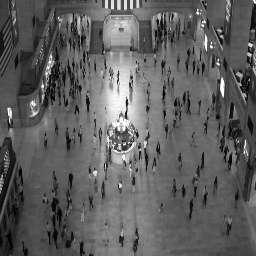}&
			\includegraphics[width=0.15\textwidth]{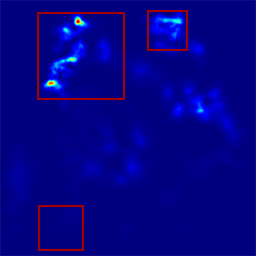}& 
			\includegraphics[width=0.15\textwidth]{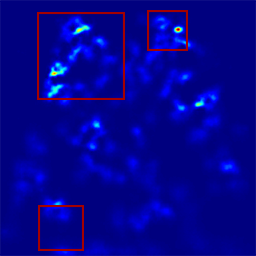}&
			\includegraphics[width=0.15\textwidth]{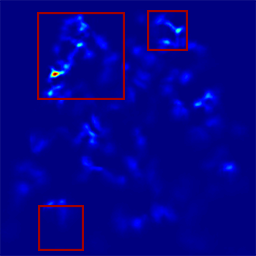}&
			\includegraphics[width=0.15\textwidth]{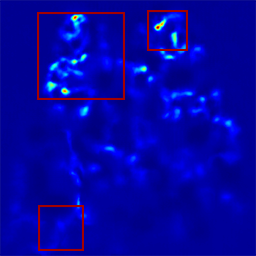}&
			\includegraphics[width=0.15\textwidth]{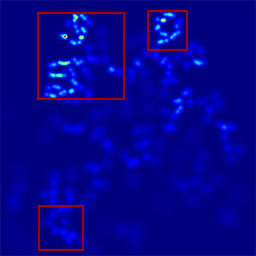}\\
			Future frame & VP & Flow + frame & Flow + density & Ours & GT density\\
			             & P-MAE: 2.041 &  P-MAE: 1.381 & P-MAE: 1.448 & P-MAE: 0.992 &  \\
			\includegraphics[width=0.15\textwidth]{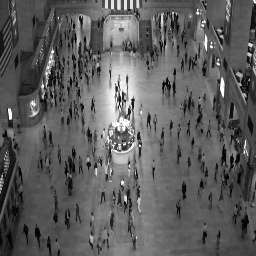}&
			\includegraphics[width=0.15\textwidth]{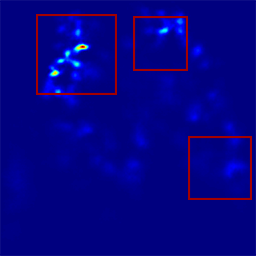}& 
			\includegraphics[width=0.15\textwidth]{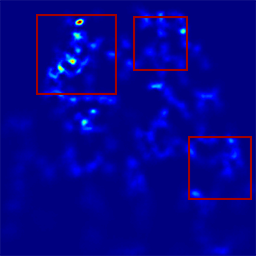}&
			\includegraphics[width=0.15\textwidth]{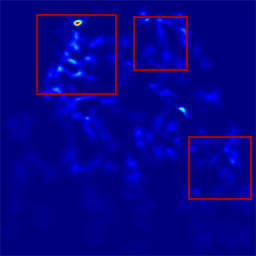}&
			\includegraphics[width=0.15\textwidth]{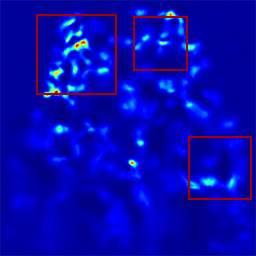}&
			\includegraphics[width=0.15\textwidth]{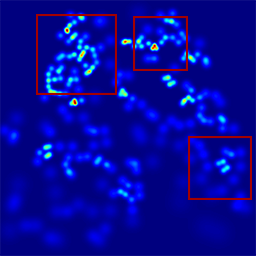}\\
			Future frame & VP & Flow + frame & Flow + density & Ours & GT density\\
			             & P-MAE: 2.232 &  P-MAE: 1.589 & P-MAE: 1.578 & P-MAE: 1.213 &  \\
			\includegraphics[width=0.15\textwidth]{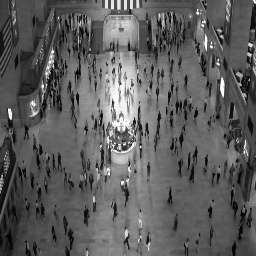}&
			\includegraphics[width=0.15\textwidth]{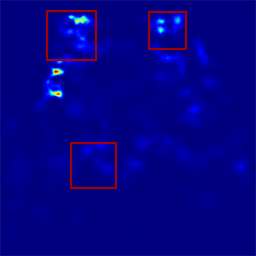}& 
			\includegraphics[width=0.15\textwidth]{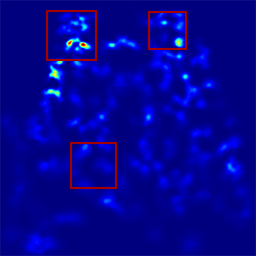}&
			\includegraphics[width=0.15\textwidth]{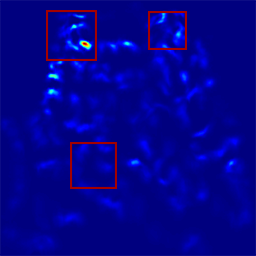}&
			\includegraphics[width=0.15\textwidth]{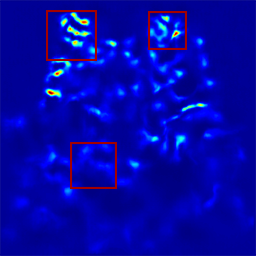}&
			\includegraphics[width=0.15\textwidth]{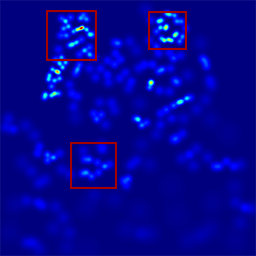}\\
			Future frame & VP & Flow + frame & Flow + density & Ours & GT density\\
			             & P-MAE: 2.063 &  P-MAE: 1.380 & P-MAE:  1.649 & P-MAE: 1.335 &  \\
		\end{tabular}
	\end{center}
	\vspace{-0.2cm}
	\caption{Examples of the comparison for crowd density prediction in the term of P-MAE$_{K=64}$.}
	\label{fig:qr_station}
\end{figure*}

\begin{figure*}
	\footnotesize
	\begin{center}
		\begin{tabular}{cccccc}
			\includegraphics[width=0.15\textwidth]{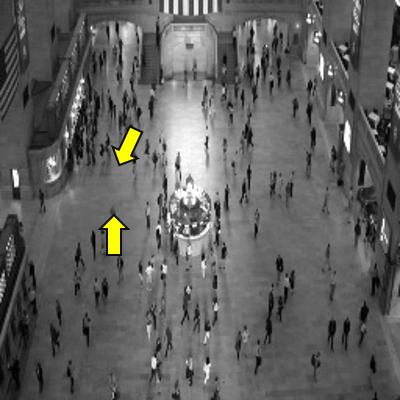}&
			\includegraphics[width=0.15\textwidth]{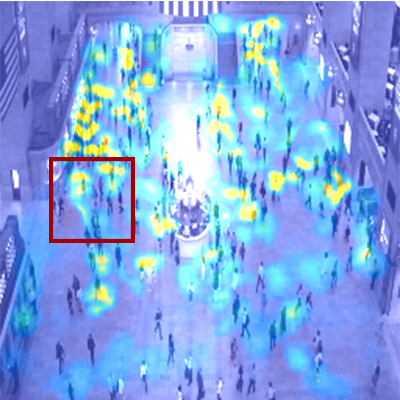}&
			\includegraphics[width=0.15\textwidth]{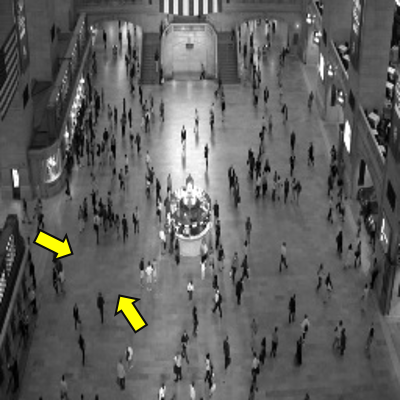}&
			\includegraphics[width=0.15\textwidth]{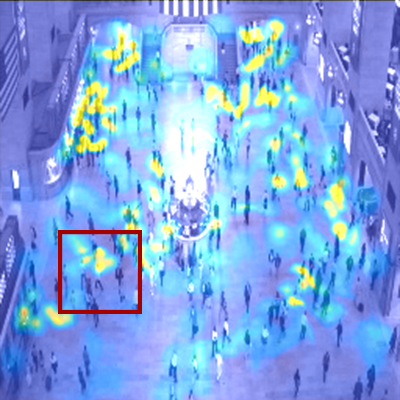}&
			\includegraphics[width=0.15\textwidth]{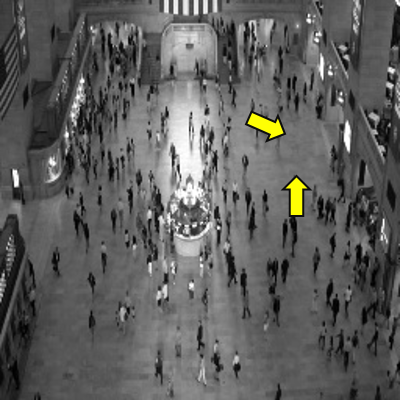}&
			\includegraphics[width=0.15\textwidth]{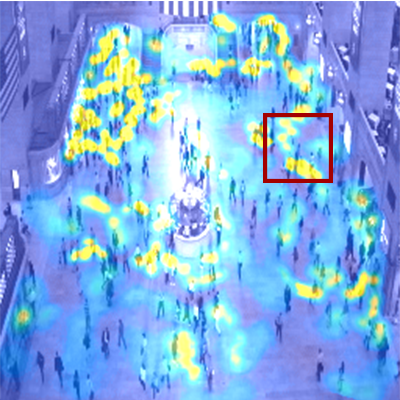}\\
			\multicolumn{2}{c}{(a)} & \multicolumn{2}{c}{(b)} & \multicolumn{2}{c}{(c)}\\
		\end{tabular}
	\end{center}
	\vspace{-0.3cm}
	\caption{We demonstrate three examples including the observed frame and the predicted high-density regions. }
	\vspace{-0.3cm}
	\label{fig:high_density}
\end{figure*}

As observed in Table~\ref{tab:cmp}, our method generally obtains better performance than the other methods. From the results of the Mall dataset, we observe that our method gains larger advantage when predicting density based on input frames with longer interval. For instance, on P-MAE$_{K=64}$, our model can gain 6.6\% and 7.8\% advantages over the second best method for $\Delta t >1.5$ and $>4.5$, respectively.
For UCSD, our approach is slightly worse than the \textit{Flow+density} method on P-MAE$_{K=64}$. 
This is probably because, as the camera view is oblique, the crowd density at the farthest end of the road is not easy to predict based on video frames. On the other hand, most pedestrians walk in straight lines for a long distance, which indicates the flow map can provide reliable information. Since its dataset duration is not too long, we cannot evaluate it on a longer interval. For Station, our metrics P-MAE$_{K=1}$ and P-MSE$_{K=1}$ are generally worse than the \textit{Flow+density} method, which implies the prediction of our crowd count for the complete scene is not well, because the scene is very large to contain hundreds of people and thus a few person entering/leaving will not affect the total count much. Thus, the warped density is adequate to estimate the total count of the scene. For more accurate estimate of crowd distribution with $K>1$, our approach demonstrates marginal advantages, especially for the prediction over longer interval (i.e. 6 seconds), which reflects our model is able to support longer-term prediction than other methods.
In Fig.~\ref{fig:qr_station}, we demonstrate examples by comparing our approach with VP~\cite{mathieu2015deep} and flow-based methods. We highlight the regions where our method provides more accurate prediction.




\subsection{High-density region prediction}

As one important application of monitoring crowd behavior, forecasting over-crowded regions can improve the crowd management and avoid danger caused by crowds. 
By learning the mapping between video frames and the future density map, our model implicitly learns the motion dynamics of crowd, and thus is able to estimate the likelihood of high-density regions due to gathering crowd. As shown in Fig.~\ref{fig:high_density}, the yellow arrows indicate the moving directions of crowd and our model accurately predicts the highlighted regions to be the high-density regions caused by the crowd gathering.

\section{Conclusion}

In this work, we formulate a novel crowd analysis problem, in which we aim to predict the crowd distribution in the near future given several consecutive crowd images. To solve this problem, we propose a global-residual two-stream recurrent network based framework, which leverages consecutive crowd frames as inputs and their corresponding density maps as auxiliary information to forecast the future crowd distribution. And we demonstrate our framework is able to predict the crowd distribution in different crowd scenarios and we delve into many crowd analysis applications. 

{\small
\bibliographystyle{ieee_fullname}
\bibliography{mooncake,crowd,traj,video}
}

\end{document}